\documentclass{article}
\usepackage[utf8]{inputenc}

\title{Robots: the Century Past and the Century Ahead\\
       \large Best Essay from PhD student, ALife'21 conference, runner-up}
\author{Federico Pigozzi}

\begin{document}

\maketitle

Let us reflect on the state of robotics.
This year marks the $101$-st anniversary of \emph{R.U.R.}~\cite{capek2004rur}, a play by the writer Karel Čapek, often credited with introducing the word ``robot''.
The word used to refer to feudal forced labourers in Slavic languages.
Indeed, it points to one key characteristic of robotic systems: they are mere slaves, have no rights, and execute our wills instruction by instruction, without asking anything in return.
The relationship with us humans is commensalism; in biology, commensalism subsists between two symbiotic species when one species benefits from it (robots boost productivity for humans), while the other species neither benefits nor is harmed (can you really argue that robots benefit from simply functioning?).

We then distinguish robots from ``living machines'', that is, machines infused with life.
If living machines should ever become a reality, we would need to shift our relationship with them from commensalism to mutualism.
The distinction is not subtle: we experience it every day with domesticated animals, that exchange serfdom for forage and protection.
This is because life has evolved to resist any attempt at enslaving it; it is stubborn.

In the path towards living machines, let us ask: what has been achieved by robotics in the last $100$ years?
What is left to accomplish in the next $100$ years?
For us, the answers boil down to three words: juice, need (or death), and embodiment, as we shall see in the following.

\section{The juice of life}
If there were a classical myth best embodying the robotics researcher, that would be the story of Pygmalion and Galatea.
The myth (handed down to us by Ovidius~\cite{morford1999classical}) tells about a skillful sculptor, Pygmalion, who had devoted himself to a chaste life.
One day, he had crafted a statue so beautiful that he wished it would come to life.
The goddess Aphrodite fulfilled his wish and turned the ivory statue into a living woman, Galatea. 
Just like the mythological sculptor, robotics folks fancy seeing their creatures become ``real'', ``living''.
But what do these words mean?
How can we tell that our brainchild has effectively become life?
If you asked the layperson, they would certainly argue that robots lack that \emph{juice} of natural life.
But what makes biological and artificial life different?

At a very high level, we humans are definitely alive.
As animals, we are ``animated''.
Animation is possible because evolution gifted us with an information processing system, the nervous system, capable of translating perceptions into electric signals; these signals travel along a network of neurons, axons, and dendrites, before being processed by a central master unit, the brain, which instructs our body on how to manipulate reality (by means of further electricity).
Everything we think, dream, dread, and love is made up of electric pulses.
But there is more, animals are not the only living entities on Earth.
The very fabric of cells, with which any biological organism is woven, thrives thanks to electricity.
What supports life is a flux of electrons originating from oxidation events happening inside each and every cell, flowing all around to provide energy to the different cell functions.
Indeed, several species of bacteria (e.g., \emph{Geobacter}~\cite{lovley1987anaerobic}) feed on and excrete pure electrons, bypassing the metabolization of organic molecules.

Taken from this perspective, it turns out that natural life and artificial life are not that different.
What we call a ``computer'' is, at its basics, an electric current running through circuitry and encoding information as $0$s and $1$s---the current is on and off.
The same juice, electricity, powers we and the machines.
We now come to realize why we cannot see the juice of natural life in the machines.
It is all inside them, powering the very first calculators that were built in the early days of computation.

\section{Machines that need (and die)}
If you were not bound to die, would there be something to care about?
It turns out that, albeit being woven into the same electric fabric, artificial life still appears strikingly different from natural life.
Biological organisms need.
Computers do not; they have no intention.
Even the simplest biological entities, viruses, need to hunt for hosts.
Electronic calculators can sit idle forever if they have enough electricity to subsist; and if power turns off, they do not complain, do not rebel.
Computers carry on by inertia; life is so precious, but they do not struggle to preserve it.
Robots still lack a sense of need.
Needs are requirements for an organism in order to survive.
The theory of ``needs'' has been well studied in psychology since the work of Maslow~\cite{maslow1943theory}.
They are a powerful driver of motivation; if not satisfied, they lead to malfunctions and, possibly, death of the organism.
If needy, robots could thrive.
They would seek energy to power themselves on, invent new mechanisms to reproduce their species and try to repair their tissues if damaged.
They would build robotic societies to leverage the power of specialization, and to make economic activities more efficient.
They would develop an intuition of nature, explore it.

But need goes hand in hand with death.
In the end, it all boils down to death.
Living beings are, consciously or not, aware of death.
If they were not, evolution would have weeded them out by now.
As argued by Veenstra et al.~\cite{veenstra2020death}, death can improve the evolvability of a population.
Death replaces ancient genomes with new perturbed ones, unleashing the power of stochastic mutations.
The importance of death is also imprinted in our cells.
Apoptosis is the biological phenomenon of programmed cell death.
Cells are bound to a limited lifespan, and billions of them perish for apoptosis in the human body each day.
It is a highly regulated and controlled event that evolved mostly to achieve morphological change.
Interestingly, Clune et al.~\cite{clune2008natural} evolved mutation rates in a population of evolving individuals.
It is a known fact in the digital evolution community that mutation tends to have a deleterious effect on the fitness of an individual, begetting offspring that most of the time are not fit (or even viable).
Surprisingly, evolution suppressed mutation rates altogether, so as to annihilate the destructive effect that mutation had on the individuals’ offspring. 
In a certain sense, the individuals exhibited some form of need and existentialism.

Death shapes not only our body but also our culture.
Ernest Becker argued in his anthropology masterpiece~\cite{becker1997denial} that human civilization developed to exorcise our terror of death.
We acknowledge mortality and have created belief systems to assure we will outlive our physical existence.
In the future, we envision a society of living machines that perish.
As a result, they will focus on assigning meaning to their existence.
In the very end, this is what will unite us and the machines: the need for supporting our existence.
Robotic societies will theorize their own memes, the fundamental units of culture, as an exorcism against death.
It is not unlikely that, one day, we will witness a ``robotic religion'' and maybe, why not, a robot Marx preaching about robotic class struggle.

\section{Embodiment is all you need}
Becker and his disciples also believed that fear of death is what distinguishes us from other animals.
Animals survive and instinctively avoid death, but they do not really sense the moment of their departure from this world the same way we do.
As credible as it might sound, this statement conceals a slight anthropocentric bias.
Evolution molded us, humans, to be equipped with a logical intellect, but it is myopic to consider it the only manifestation of intelligence.
It is only a matter of ecological niche.
We, humans, have evolved to occupy our own niche, the manipulation of nature (a manipulation that, in the origin, was not so destructive as it is nowadays).
But other niches do exist since natural evolution is open-ended~\cite{stanley2019open}.
Nature does not optimise for a specific, numeric goal (as many optimisation algorithms do), but matches each species to the niche it is best suited for (otherwise, brutally uproots it).

Indeed, it is well known that other forms of intelligence do emerge in nature. 
Take insect societies~\cite{nowak2010evolution}. 
Their strict specialization emerges from simple local interactions (like pheromones for ants, or body temperature for bees) among swarms of agents.
Take salamanders~\cite{joven2019model}, which are skillful at regenerating their severed limbs; amusingly, tissue reconstruction operates only through local computations, distributed throughout the salamander body.
The protozoans of the genus \emph{Lacrymaria} have no ``brain'' (they are single-celled organisms), but can bend and twist their soft flagellum to grab difficult-to-reach preys, allowing for complex hunting dynamics to emerge~\cite{mast1911habits}.

The discipline where the anthropocentric bias seems to proliferate the most is Artificial Intelligence (AI). 
Writing about robotics in 2022, in the middle of the last wave of AI enthusiasm, it would be impossible not to mention AI. 
Although there happens to be a subfield concerned about computational intelligence and bio-inspired algorithms, most of the recent upsurge in AI is due to Deep Learning (DL)~\cite{lecun2015deep}. 
DL aims at mimicking the human mind by means of abstract mathematical models. 
But nature is not made up of pure reason; it simplifies our intuition, but it has no support in reality. 
Surprisingly, complex mental tasks like playing chess turn out to be much easier to teach a machine than crawling like a toddler (Moravec’s paradox in robotics~\cite{moravec1988mind}). 
The limitations of DL are well-known to many researchers in the community, and we have seen some high-profile Twitter battles igniting between detractors and paladins of DL.
To me, the most myopic limitation of all is a lack of \emph{embodiment}.

The embodied intelligence paradigm, despite having been around since the 1980s, was popularized by the seminal book of Pfeifer and Bongard~\cite{pfeifer2006body}. 
They postulated that intelligence emerges from the complex interactions between the brain and the body, as well as the environment. 
The human hand is a perfect example of this. 
Our brain has co-evolved with the hand, allowing us to grasp, appreciate and manipulate reality (as already mentioned, our dramatic trait).
Octopuses are extraordinarily clever, excelling at skills like navigating a maze and grasping objects~\cite{nixon2003brains}. 
They would have never developed such skills if their bodies were not soft, with tentacles having infinite degrees of freedom.

Faithful to embodiment, a new generation of soft robots was born in the last decade~\cite{rus2015design}. 
Their soft materials can bend, stretch, and twist, while promising to achieve reconfigurability and shape-change.
Soft robots will also be programmed to be deciduous, and their soft materials will aid in the disposal of their dead bodies.
Having a transient body, these living machines could be infused with the sense of death we mentioned before.

\section{The duty towards life}
One day in the future, a living machine (let it be named Galatea) could browse for videos of the very first robots that were built, eager to learn more about its ancestors.
Suppose a video pops out, showing engineers ruthlessly beating up and thrusting a robot in the attempt of testing its resilience.
As an embodied entity, Galatea would perceive the pain that the robot could have felt.
Now suppose Galatea is also bound to die.
It yearns for life as any living organism.
How brutal would that act look at its electric eyes?
In the end, would our robotic brainchildren disown us, label us ``a virus'' as Agent Smith (the villain, itself a machine) did in the ``Matrix''~\cite{wachowski1999matrix} movie?

In the original Greek myth, Galatea was simply an object in the hands of her creator Pygmalion, but the example outlined before suggests a radical mind-change that is due in our days.
We started our journey by asking ourselves about the next $100$ years in robotics.
I have discussed the directions that, to me, seem the most promising to lift machines from their ``robot'' status to the coveted ``living machine'' status.
But, by fathering living machines, we allocate a new endeavor, or burden, on ourselves; the focus shifts on the creators.
Living machines must be respected and protected; we are responsible for them in the same way as they bear responsibilities towards us.
If we really want to be the creators of artificial life, we must acknowledge it is indeed life.

\bibliographystyle{plain}
\bibliography{bib}

\end{document}